\pdfoutput=1
\documentclass[10pt,twocolumn,letterpaper]{article}

\usepackage{cvpr}
\usepackage{times}
\usepackage{epsfig}
\usepackage{graphicx}
\usepackage{amsmath}
\usepackage{amssymb}
\usepackage{subcaption}
\usepackage{multirow}
\usepackage[table]{xcolor}
\usepackage{color}
\usepackage{colortbl}
\usepackage{tabularx}
\usepackage{bm}
\usepackage{booktabs}
\usepackage{url}
\usepackage{stackengine}
\usepackage{float}
\usepackage{verbatim}

\usepackage[breaklinks=true,bookmarks=false]{hyperref}

\cvprfinalcopy 

\def\cvprPaperID{****} 

\begin{document}

\title{Improving Semantic Segmentation of Aerial Images Using Patch-based Attention}

\author{Lei Ding\\
University of Trento\\
Via Sommarive 5, Trento, Italy\\
{\tt\small lei.ding@unitn.it}
\and
Hao Tang\\
University of Trento\\
Via Sommarive 5, Trento, Italy\\
{\tt\small hao.tang@unitn.it}
\and
Lorenzo Bruzzone\\
University of Trento\\
Via Sommarive 5, Trento, Italy\\
{\tt\small lorenzo.bruzzone@unitn.it; bruzzone@ieee.org}
}
\maketitle

\begin{abstract}
    The trade-off between feature representation power and spatial localization accuracy is crucial for the dense classification/semantic segmentation of aerial images. High-level features extracted from the late layers of a neural network are rich in semantic information, yet have blurred spatial details; low-level features extracted from the early layers of a network contain more pixel-level information, but are isolated and noisy. It is therefore difficult to bridge the gap between high and low-level features due to their difference in terms of physical information content and spatial distribution. In this work, we contribute to solve this problem by enhancing the feature representation in two ways. On the one hand, a patch attention module (PAM) is proposed to enhance the embedding of context information based on a patch-wise calculation of local attention. On the other hand, an attention embedding module (AEM) is proposed to enrich the semantic information of low-level features by embedding local focus from high-level features. Both of the proposed modules are light-weight and can be applied to process the extracted features of convolutional neural networks (CNNs). Experiments show that, by integrating the proposed modules into the baseline Fully Convolutional Network (FCN), the resulting local attention network (LANet) greatly improves the performance over the baseline and outperforms other attention based methods on two aerial image datasets.
\end{abstract}

\section{Introduction}\label{Sec1}
\begin{figure}[t]
\centering
    \includegraphics[width=1\linewidth]{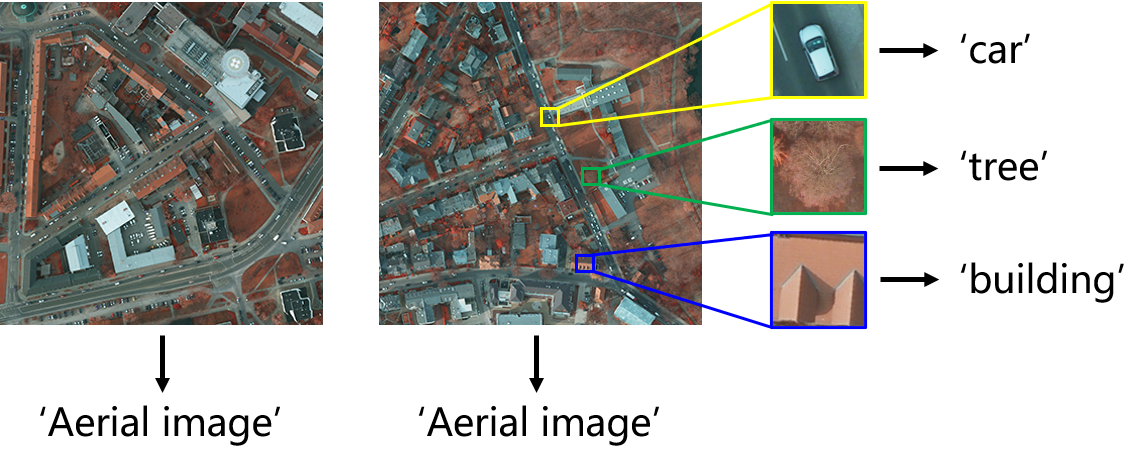}
    \caption{Examples of the image-level information for aerial images. The information of a whole aerial image cannot be deduced more specifically than just `aerial image', but the information of image patches can be easily attributed to classes like `car, `tree' and `building'.}
\label{img_info}
\end{figure}

Images collected from aerial platforms are widely used in a variety of applications, such as land-use mapping, urban resources management and disaster monitoring. Semantic segmentation, namely the pixel-wise classification of images, is a crucial step for the automatic analysis and explanation in applications of these aerial data. The rise of convolutional neural networks (CNNs) and the emergence of Fully Convolutional Networks (FCNs)~\cite{long2015fcn} has brought a breakthrough in semantic segmentation of aerial images~\cite{audebert2018beyondrgb}. Typical CNN architectures used in visual recognition tasks employ cascade spatial-reduction operations to force the networks to learn intrinsic representations of the observed objects~\cite{he2016resnet}. However, this so-called `encoding' design has the side-effect of losing spatial information. The classification maps produced by encoding networks usually suffer a loss of localization accuracy (e.g., the boundaries of classified objects are blurred and some small targets may be neglected). Although there are `decoding' designs to recover spatial information by using features extracted from early layers of the CNNs~\cite{badrinarayanan2017segnet,ronneberger2015unet,lin2017refinenet}, their effectiveness is limited due to the gap between the high-level and low-level features in both semantic information and spatial distribution~\cite{zhang2018exfuse}.

This trade-off between feature embedding power and spatial localization accuracy is crucial for the semantic segmentation of aerial images. On the one hand, different categories of the ground objects may share similar spectral features in aerial images, thus requiring for an aggregation of the context information. On the other hand, many applications of analysing aerial images require high precision in mapping contours of ground objects. Therefore, detailed spatial information is needed for identifying accurately both the boundary of regions and small objects.

The introduction of attention mechanism is an effective strategy to reduce the confusion in predicted categories without losing spatial information. With the global statistics aggregated from the whole image, scene information can be embedded to highlight (or suppress) the features with strong correlations~\cite{hu2018squeeze}. However, the spatial size of aerial images is usually much larger than that of natural images, whereas the number of object categories is smaller. For example, each image in the ISPRS semantic labelling dataset (Potsdam area) ~\cite{isprs2Dlabelling} has 6000 $\times$ 6000 pixels divided into 6 object categories in this dataset. As a result, almost every image contains all the object categories, and no clear global scene information can be embedded at the global level. In other words, we argue that the typical attention-based techniques cannot be directly applied to the semantic segmentation of large-size aerial images.

In this paper, we propose the generation of patch-level attention to improve the semantic segmentation of aerial images. The proposed approach is based on the finding that, although the semantic information of a whole aerial image cannot be specifically concluded, the image patches still have clear semantic reference (an illustration example of this observation is given in Fig.~\ref{img_info}). Therefore, we propose a novel Patch Attention Module (PAM) to exploit patch-wise local attention. This module operates on extracted feature maps and can aggregate context information from the local patch to reduce confusions. In our model, the PAM is appended after both the high-level and low-level features to enhance their representation. Moreover, to bridge the gap between high-level and low-level features, an Attention Embedding Module (AEM) is proposed to embed semantic focus from high-level features to low-level features. This module can greatly improve the semantic representation of low-level features without losing their spatial details, thus improving the effectiveness of the fusion between high-level and low-level features. The proposed modules are light-weight and can be incorporated into existing CNN architectures to improve the classification accuracy. Using a FCN as baseline network, we performed experiments on two aerial datasets and proved the the effectiveness of the proposed techniques.

Our contributions can be summarized as follows:
\begin{itemize}
\item Proposing a local attention network (LANet) to improve the semantic segmentation of aerial images by enhancing the scene-related representation in both encoding and decoding phases.
\item Proposing both patch attention module (PAM) to embed scene information from local patches, and attention embedding module (AEM) to enhance the semantic representation of low-level features by introducing attention from high-level features.
\item Extensive ablation studies have been performed by incorporating the proposed modules into the baseline FCN network in sequence. The resulting LANet is further compared with other networks with decoding or attention-based designs to evaluate its performance.
\end{itemize}

\section{Related Work}\label{sc2}

\begin{figure*}[t]
    \centering
    \includegraphics[width=1\linewidth]{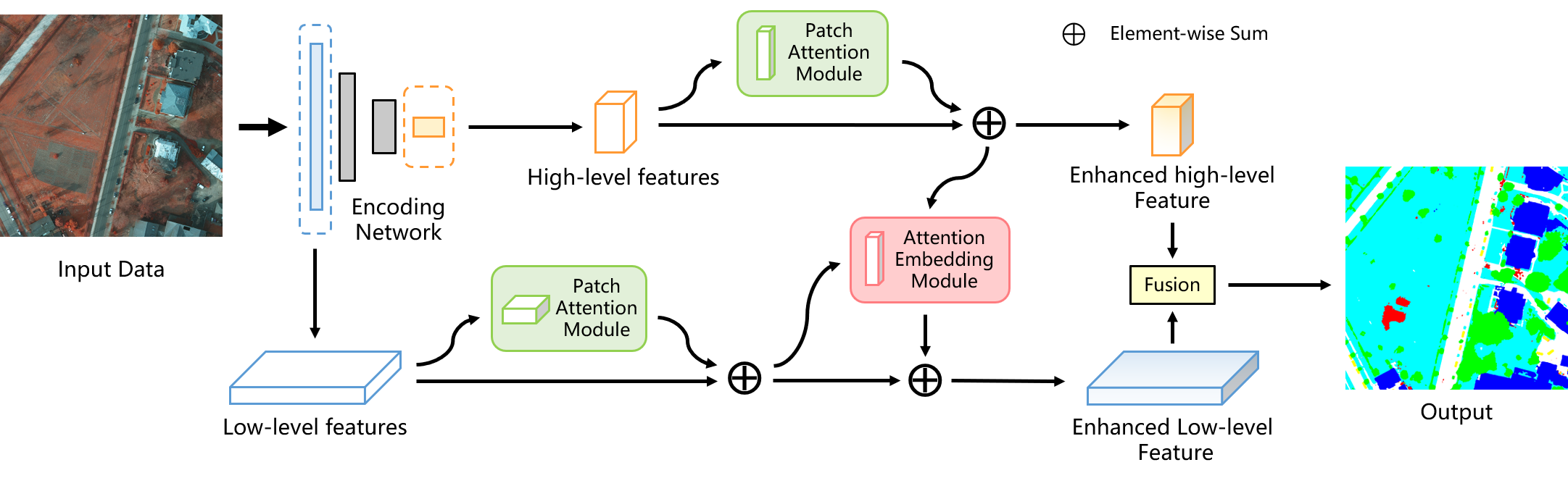}
    \caption{Architecture of the proposed local attention network (LANet). The patch attention module (PAM) generates attention maps to highlight patch-wise focus in feature maps. The attention embedding module (AEM) embeds semantic information from high-level features to low-level ones.}
    \label{overview}
\end{figure*}

\subsection{Semantic Segmentation of Aerial Images}
Semantic segmentation on aerial scenes has drawn great research interests after the rising of CNNs and the publish of several open datasets/contests such as ISPRS Benchmarks\footnote{http://www2.isprs.org/commissions/comm3/wg4/semantic-labeling.html}, DeepGlobe contest\footnote{http://deepglobe.org/challenge.html}, and SpaceNet competition\footnote{https://spacenetchallenge.github.io/}.  Several studies incorporate multiple models to increase the prediction certainty \cite{paisitkriangkrai2015CNNhc,zhang2018cnnmrf,yu2018PSPResnet}. The prediction of object contours is an issue of concern. Detection of edges is explicitly added in \cite{marmanis2018boundarydetection}, while \cite{liu2018edgeloss} introduced an edge loss to enhance the preservation of objects boundaries. The utilization of other forms of data (e.g., Lidar data, digital surface models and OpenStreetMap) is also widely studied  \cite{kaiser2017osmpretrain,sun2018DSMrefine,sun2018featureFus,audebert2017OSMinput}. However, there are limited studies focused on the special characteristics of aerial images (e.g. large spatial size, fixed imaging angle and small number of classes). In this work we consider these characteristics when designing the specific modules.

\subsection{Encoder-Decoder Designs}
The encoder-decoder networks have been successfully used in many computer vision tasks such as image generation~\cite{isola2017image,tang2019multi}, object/saliency detection~\cite{lin2017feature,min2019tased}, crowd counting~\cite{jiang2019crowd} and semantic segmentation~\cite{chen2018encoder,long2015fully}.
Usually, the encoder-decoder networks contain two sub-nets:
(i) an encoder sub-net that gradually reduces the feature maps and captures higher semantic information,
and (ii) a decoder sub-net that gradually recovers the spatial information. The encoder sub-net is the focus of most existing studies. There are many works related to enlarging the receptive field without significantly increasing the number of parameters ~\cite{yu2018PSPResnet,chen2018deeplabv3+}. Although in some studies there are cascade decoding designs that aim to exploit the features from early CNN layers ~\cite{ronneberger2015unet,badrinarayanan2017segnet,lin2017refinenet,zhang2018exfuse}, these features are usually concatenated or summed to the high-level features without enhancing their semantic representation. Thus, they provide limited contribution to the classification accuracy. To overcome this limitation, we propose the use of attention mechanism for enhancing the representation of low-level features during the decoding phase.

\subsection{Attention Mechanism}
Attention mechanism refers to the strategy of allocating biased computational resources to the processed signal to highlight its informative parts. In the tasks related to the understanding of image content, a typical solution for generating attention statistics is to gather information from a global scale, namely to exploit the scene or image-level information. This is because the scene information may provide clues about the possible contents in an image. In~\cite{wang2017residual}, the attention of the feature map is aggregated using an hourglass module in a residual manner. This residual attention network introduced a chunk-and-mask module, where the global attention is aggregated in the Soft Mask Branch through stacked down-sampling convolutions. In~\cite{hu2018squeeze}, a Squeeze-and-Excitation (SE) block is proposed, which uses global-pooling to generate channel-wise attention. In this way, spatial-irrelevant information can be learned to emphasize the scene-relevant feature channels. The design of ‘squeezing’ spatial information and the parallel connection of attention branch introduced in this work have been widely adopted in subsequent studies. In EncNet~\cite{zhang2018context}, a context encoding module is proposed to capture the scene-dependent global context as channel-wise attention. CBAM~\cite{woo2018cbam} introduced a spatial attention module to highlight the informative spatial regions. The spatial attention maps are generated by using pooling operations along the channel axis. BAM~\cite{park2018bam} has a similar module to exploit spatial correlations but it is implemented by applying dilated convolutions. PSANet~\cite{zhao2018psanet} introduced the modelling of long-range correlation for each spatial position, but the channels of its inner layers are related to the input image size and cannot be applied to the prediction of full-size aerial images. A parallel design that models both channel-wise and point-wise attention is introduced in DANet~\cite{nam2017dual}. A limitation of this network is that the reasoning of global spatial correlation is calculation intensive. A light-weight graph-based module for reasoning latent correlations has been presented in~\cite{chen2019glore}.

Building on top of these studies, we propose a simple yet effective approach that extends the use of attention mechanism to the spatial dimension without significantly increasing the computational load.

\section{Proposed Approach}\label{sc3}
In this section we present the proposed LANet devised for improving semantic segmentation of aerial images. Firstly, an overview of the network is given to introduce the general  motivation and architecture. After that, the proposed modules are described in detail. Finally, a further explanation on aggregation strategy for different levels of features is given.

\subsection{Overview of the Proposed LANet}
The motivation of this work is to strengthen the representation of features extracted from backbone CNNs while minimizing the loss of spatial details. To achieve this goal, we propose a LANet with two separate modules: a patch attention module (PAM) to enhance the embedding of local context information, and an attention embedding module (AEM) to improve the use of spatial information. Specifically, we designed two parallel branches to process features from different layers. As shown in Fig.~\ref{overview}, in the upper branch, high-level features (produced by late layers of a CNN) go through a PAM to enhance their feature representation; in the lower branch, low-level features (produced by early layers of a CNN) are first enhanced by PAM, then embedded with semantic information from high-level through AEM. The final classification results are produced by the fusion of the features from both branches.

\subsection{Patch Attention Module}
\label{approach_PAM}

\begin{center}
    \begin{figure}[tbbp]
    \begin{center}
    {\includegraphics[width=1\linewidth]{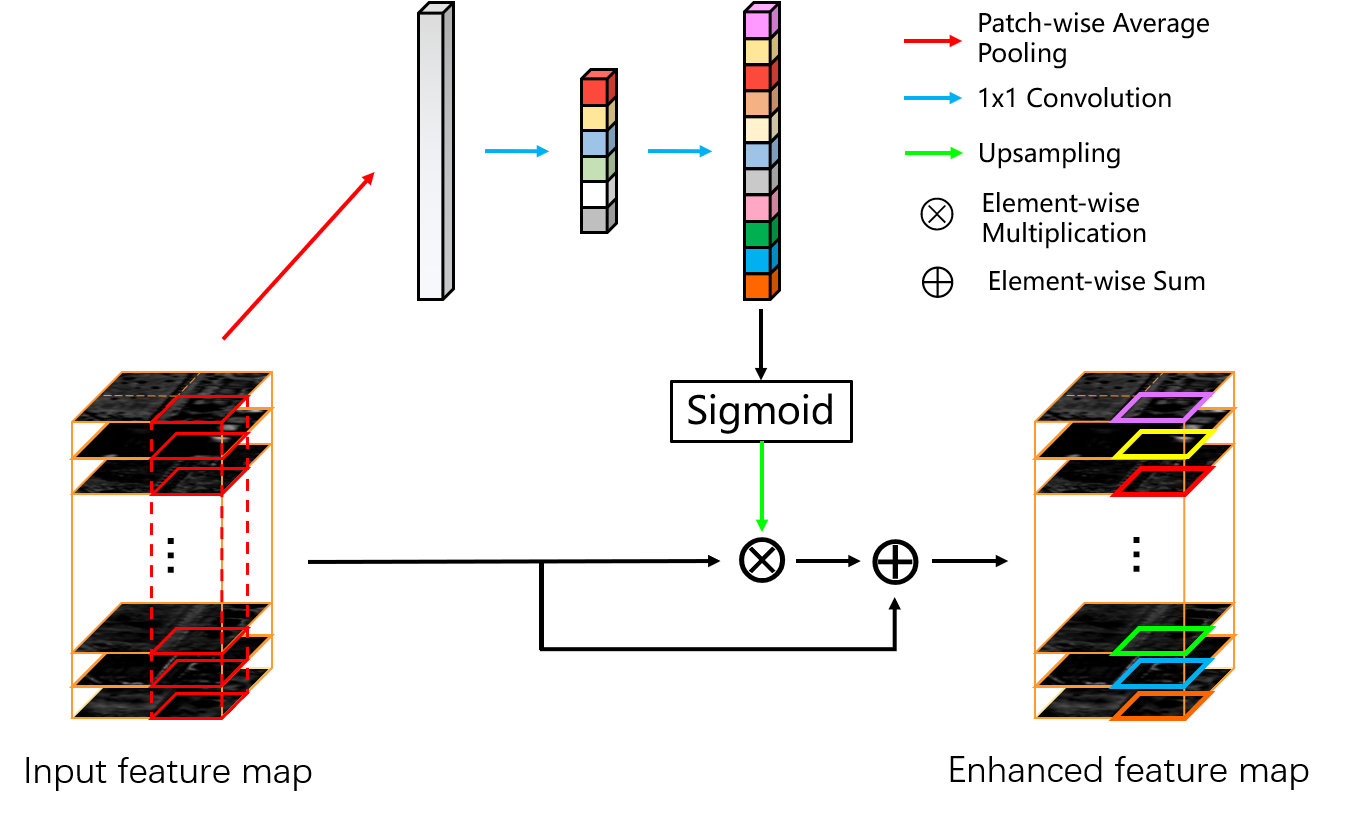}}
    \end{center}
    \caption{Detailed design of the PAM. Descriptors are calculated patch-wisely to aggregate local context information.}\label{PAM}
    \end{figure}
\end{center}

Semantic segmentation of aerial images suffers greatly from the problem of intra-class inconsistence, since the classification of ground objects is a comprehensive task affected by both the surface type and the context of an image. To alleviate this problem, we propose a patch attention module to enhance the aggregation of context information in the extracted features.

Fig.~\ref{PAM} shows the design of the PAM. Our work is inspired by the design of the SE-block~\cite{hu2018squeeze}. The original SE-block introduced global average pooling to generate one single descriptor for each feature channel. However, as discussed in Section~\ref{Sec1}, this cannot be applied to the processing of large-size aerial images. In our approach, we limit the generation of descriptors in patches, so that each descriptor contains meaningful information of the local context. The descriptor $z_c$  for the c-th channel of a patch is calculated as:
\begin{equation}\label{equation_descriptor}
z_c = \frac{1}{h_{p} \times w_{p}} \sum_{i=1}^{h_{p}}\sum_{j=1}^{w_{p}}x_{c}(i,j),
\end{equation}
where $h_{p}$ and $w_{p}$ denote the spatial size of the pooling window, $x_{c}$ denotes a pixel at $c$th channel. In this way, a $c$-channel vector $\textbf{z}_p$ can be generated, which contains the statistics describing the patch $p$.
After that, we follow the bottleneck gating design in~\cite{hu2018squeeze} to learn an attention vector $\textbf{a}_{p} \in \mathbb{R}^{c \times h_{p} \times w_{p}}$ for the patch $p$. Instead of using fully connected layers, we employ convolutional operations so that they can be applied to process other patches without assigning extra weights. The gating operation to generate attention maps can be symbolized as:
\begin{equation}
\textbf{a}_{p} =  \textbf{F}_{U}  \{ \sigma [ H_{i} \delta (H_{r}\textbf{z}_{p}) ] \},
\end{equation}
where $\sigma$ and $\delta$ denote Sigmoid and ReLU functions~\cite{nair2010relu}, respectively; $H_{r}$ denotes the 1$\times$1 dimension-reduction convolution with the reduction ratio $r$, $H_{i}$ denotes the 1$\times$1 dimension-increasing convolution that recovers the feature dimension back to $c$. $\textbf{F}_{U}$ is the upsampling operation.

This is the case for a single local patch. Now we consider it at the global level. Given a feature map $\textbf{X} \in \mathbb{R}^{C \times H \times W}$, maps of descriptors $\textbf{Z} \in \mathbb{R}^{C \times H' \times W'}$ can be generated. $H'$ and $W'$ are determined by the size of each patch (pooling window) as:
\begin{equation}
H' = \frac{H}{h_{p}} , W' = \frac{W}{w_{p}},
\end{equation}
where $h_p$ and $w_p$ are set according to the spatial reduction ratio of the corresponding encoding layer to ensure a remarkable enlargement of the receptive field. An alternative is to use a sliding window for generating the descriptors, so that the descriptor maps have the same size of input images. However, this option will tremendously increase the calculation, thus, it is not adopted in our implementation. After the convolutional layers, attention maps $\textbf{A} \in \mathbb{R}^{C \times H \times W}$ can be produced. Finally, the original input features $\textbf{X}$ are multiplied element-wisely with $\textbf{A}$ to enhance their representation. A residual design is adopted to ensure the stable back-propagation of gradients.

\begin{center}
    \begin{figure}[tbbp]
    \begin{center}
    {\includegraphics[width=1\linewidth]{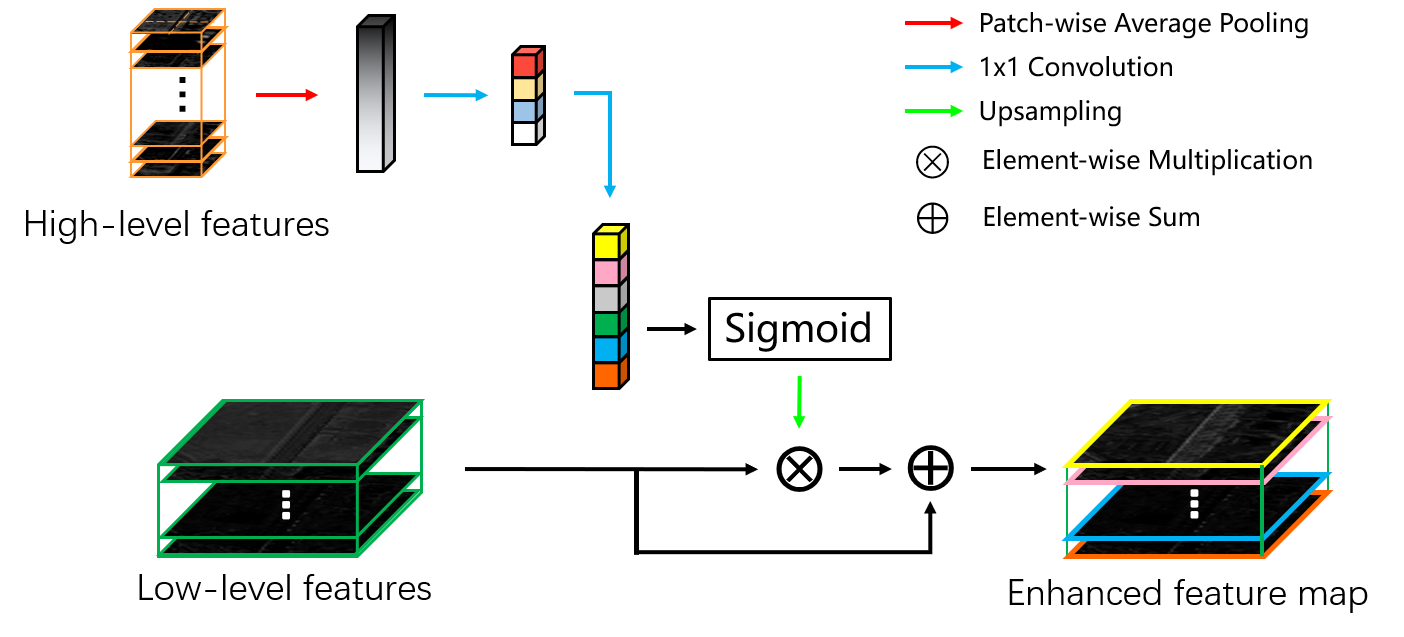}}
    \end{center}
    \caption{Detailed design of the AEM. Low-level features are semantically enriched by embedding local focus from high-level features.}\label{AEM}
    \end{figure}
\end{center}

\subsection{Attention Embedding Module}
\label{approach_AEM}
An effective exploitation of low-level features is difficult due to their difference with high-level features in terms of spatial distribution and physical meaning. The most frequently used way of employing low-level features is to concatenate them with high-level features, which brings only slight improvement in performance (refer to discussion in Section~\ref{Sec4}). To make the best use of low-level features, we propose an attention embedding module to enrich their semantic meaning. This operation bridges the gap between high-level and low-level features without sacrificing the spatial details of the latter.

Fig.~\ref{AEM} shows the design of the proposed AEM. The intuition of this approach is to embed local attention from high-level features into the low-level features. In this way, low-level features are embedded with context information that goes beyond the limitation of their receptive fields, while their spatial details are kept. First, we generate descriptors from high-level features through the same calculation as in Eq.~\eqref{equation_descriptor}. Denote these maps of descriptors as $\textbf{Z}_{h} \in \mathbb{R}^{C_{h} \times H' \times W'}$, and the low-level features as $\textbf{X}_{l} \in \mathbb{R}^{C_{l} \times H_{l} \times W_{l}}$. We generate attention maps for the low-level features $\textbf{A}_{l}$ by transforming $\textbf{Z}_{h}$ through bottleneck convolutions as:
\begin{equation}
    \textbf{A}_{l} =  \textbf{F}_{U} \{\sigma [H_{l} \delta (H_{r}\textbf{Z}_{h}) ] \},
\end{equation}
where $H_{r}$ is a dimension reduction convolution and $H_{l}$ changes the number of channels to be the same as $\textbf{X}_{l}$. To avoid excessive interference of high-level features, we add a residual design to emphasize the importance of low-level features. The enhanced low-level features are calculated as:
\begin{equation}
    \textbf{X}_{l} = \textbf{X}_{l} + \textbf{X}_{l}\textbf{A}_{l}
\end{equation}

\subsection{Feature Fusion between Different Layers}\label{approach_AEM}
After being processed by AEM, low-level features are semantically enriched and can potentially give a higher contribution to the prediction of the pixel class. Both the high-level and low-level features keep their dimensions after the processing of PAM and AEM. Accordingly, classic feature fusion operations (e.g., concatenation) can be applied to the outputs of the two branches. Since the specific feature fusion operation is not the focus of this work, also considering the convenience of validating the output from each branch, we simply train two separate classifiers for each branch, and perform an element-wise sum to generate the final results.

\section{Experiments}\label{sc4}

To assess the effectiveness of the proposed method, experiments have been conducted on two aerial image datasets, i.e, the Potsdam dataset and the Vaihingen dataset. First we provide a short description of both datasets and implementation details. Then we test the proposed modules through an ablation study. Finally, we compare the proposed LANet with state-of-the-art methods and draw the conclusion of our experimental validation.
\subsection{Experimental Setting}
\noindent \textbf{Datasets.} We employ two public available datasets to evaluate the proposed methods.

(i) The potsdam dataset~\cite{isprs2Dlabelling} consists of 38 TOP tiles and the corresponding DSMs collected from a historic city with large building blocks. 24 imageries are used for training and the remaining 14 for testing. There are four spectral bands in each TOP (red, green, blue and near infrared) and one band in each DSM. All data files have the same spatial size, equal to 6000 $\times$ 6000 pixels. The ground sampling distance (GSD) of this dataset is 5cm. The reference data are labeled according to six land-cover types: impervious surfaces, building, low vegetation, tree, car and clutter/background.

(ii) The vaihingen dataset~\cite{isprs2Dlabelling} contains 33 true orthophoto (TOP) tiles and the corresponding digital surface models (DSMs) collected from a small village. 16 imageries are used for training and the remaining 17 ones for testing. Different from the Potsdam dataset, each TOP in the Vaihingen dataset contains three spectral bands (near infrared, red and green bands) and one DSM band. The spatial size of the images varies from 1996 $\times$ 1995 pixels to 3816 $\times$ 2550 pixels. The GSD of this dataset is 9 cm. The reference data are divided into the same six categories as the Potsdam dataset.

\noindent \textbf{Evaluation Metrics.}
Following the evaluation method provided by the data publisher~\cite{isprs2Dlabelling} and used in literature~\cite{marmanis2018boundarydetection,yu2018PSPResnet,liu2017hourglass}, three evaluation metrics are used to evaluate the performance of methods, i.e, overall accuracy (OA), per-class F1 score and average F1 score. OA is calculated by dividing the correctly classified number of pixels with the total number of pixels. The F1 score for a certain class is defined as the harmonic mean of precision and recall:
    \begin{equation}
    \rm F1=2\times{\frac{precision\times recall}{precision+recall}}
    \label{F1}
    \end{equation}
    
\noindent \textbf{Implementation Settings.}
The same preprocessing, data augmentation and weight initialization settings have been used in all the experiments. The DSMs are concatenated with TOPs as input data, so that we obtain five channels for the Potsdam dataset and four channels for the Vaihingen dataset. Due to the limitation of computational resources, the input data are cropped using a 512 $\times$ 512 window during the training phase. However, the prediction for the test set is performed whole-image-wise to obtain an accurate evaluation of the compared methods. Random-flipping and random-cropping operations are conducted during each iteration of the training phase as an augmentation approach. We use ResNet50 as the backbones for all compared networks with the pretrained weight for Pascal VOC dataset loaded from the PyTorch library. Considering the different GSD of the two datasets, the down-sampling stride for the Potsdam dataset is set to 32, while for the Vaihingen dataset it is set to 16. The networks are implemented with PyTorch and the experiments are conducted on a server with a NVIDIA Quadro P6000 23GB GPU.

\subsection{Experimental Results}

\noindent \textbf{Ablation Study.}
In order to verify the effectiveness of the proposed modules, ablation studies have been conducted on the two datasets. FCN (ResNet-50) is used as the baseline network for comparison. Since the proposed LANet uses low-level features, the effect of considering low-level features has also been measured.

Table~\ref{Ablation_PD} shows the results of the ablation study on the Potsdam dataset. Three groups of observations can be done from the results. When no low-level features are involved in the decoding stage, the use of only one PAM (added on top of the FCN) increases the OA of 0.19$\%$. With the inclusion of low-level features (concatenated with high-level features), the OA of the baseline FCN increases of only 0.16$\%$. However, when two PAMs are added to process the high-level and low-level features separately, the OA increases of another 1.07$\%$. When the proposed AEM is used instead to enhance low-level features, the OA increases of 1.02$\%$. With the use of both PAM and AEM, the proposed LANet increases the OA and average F1 compared with the baseline FCN (with the use of low-level features) of 1.26$\%$ and 0.72$\%$, respectively.

The ablation study on the Vaihingen dataset is presented in Table~\ref{Ablation_VH}. Under the condition that low-level features are considered, the proposed LANet improves the average F1 score and OA of 1.57$\%$ and 0.99$\%$, respectively.

\begin{table}[tbp]
	\centering
    \caption{Results of the ablation study on the Potsdam dataset. ($^{*}$) low-feat indicates the use of low-level features. }
   \resizebox{1\linewidth}{!}{%
        \begin{tabular}{l|c|cc|cc}
        \toprule
            Method & low-feat$^{*}$ & PAM & AEM & mean F1 & OA\\
            \hline
            FCN    &  &  &  & 88.66 & 89.42 \\
            FCN+PAM & & $\surd$ &  & 89.03 & 89.61 \\
            \hline
            FCN & $\surd$ &  &  & 91.23 & 89.58 \\
            FCN+PAM & $\surd$ & $\surd$ &  & 91.76 & 90.65 \\
            FCN+AEM & $\surd$  &  & $\surd$ & 91.78 & 90.60 \\
            \hline
            LANet & $\surd$ & $\surd$ & $\surd$ & \textbf{91.95} & \textbf{90.84}\\
        \bottomrule
        \end{tabular}}
        \label{Ablation_PD}
\end{table}

\begin{table}[t]
\centering
    \caption{Results of the ablation study on the Vaihingen dataset.}
    \resizebox{1\linewidth}{!}{%
        \begin{tabular}{l|c|cc|cc}
        \toprule
            Method & low-feat & PAM & AEM & mean F1 & OA\\
            \hline
            FCN    &  &  &  &  86.14 & 88.66 \\
            FCN+PAM &  & $\surd$ &  & 86.42 & 88.68 \\
            \hline
            FCN & $\surd$ &  &  & 86.52 & 88.84 \\
            FCN+PAM &  $\surd$ & $\surd$ &  & 87.49 & 89.36 \\
            FCN+AEM & $\surd$  &  & $\surd$ & 86.80 & 89.05 \\
            \hline
            LANet & $\surd$ & $\surd$ & $\surd$ & \textbf{88.09} & \textbf{89.83}\\
        \bottomrule
        \end{tabular}}
        \label{Ablation_VH}
\end{table}

\begin{figure}[t]
\centering
    {\includegraphics[width=8cm]{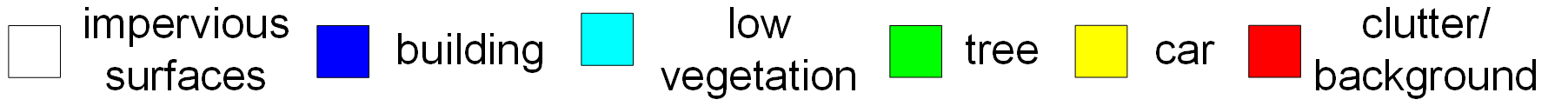}}\\
    {\includegraphics[width=2cm]{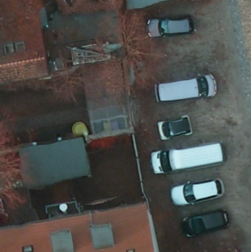}}
    {\includegraphics[width=2cm]{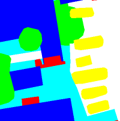}}
    {\includegraphics[width=2cm]{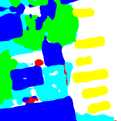}}
    {\includegraphics[width=2cm]{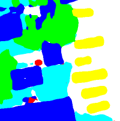}}\\
    \footnotesize
    \stackunder[5pt]{\includegraphics[width=2cm]{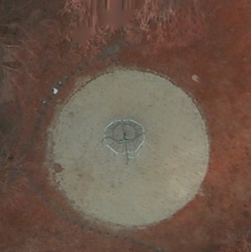}}{Image(IRRG)}
    \stackunder[5pt]{\includegraphics[width=2cm]{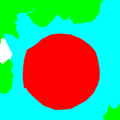}}{Reference}
    \stackunder[5pt]{\includegraphics[width=2cm]{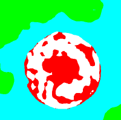}}{Before PAM}
    \stackunder[5pt]{\includegraphics[width=2cm]{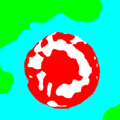}}{After PAM}
    \caption{Comparison of classified high-level features before and after the use of PAM (the Potddam dataset).}\label{Reslt_change_high}
\end{figure}

\noindent \textbf{Visualization of Features.}
To visually confirm the effect of the proposed modules, we present comparisons of the classified features generated independently before and after the use of the proposed modules. Fig.~\ref{Reslt_change_high} shows the effect of the PAM module for high-level features. Since high-level layers already have relatively large receptive field before using PAM, the enhancement is not significant. However, one can still observe that some of the meaningless small areas are removed, and the large objects become more complete. 

\begin{figure}[t]
\centering
    {\includegraphics[width=8cm]{ColorBar_h.png}}\\
    {\includegraphics[width=1.5cm]{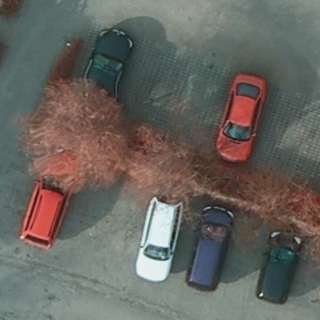}}
    {\includegraphics[width=1.5cm]{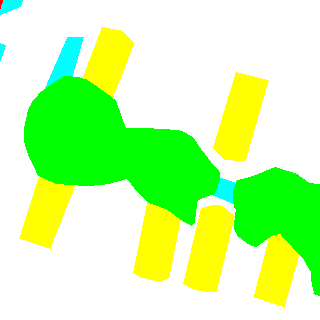}}
    {\includegraphics[width=1.5cm]{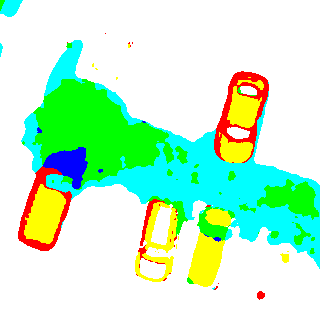}}
    {\includegraphics[width=1.5cm]{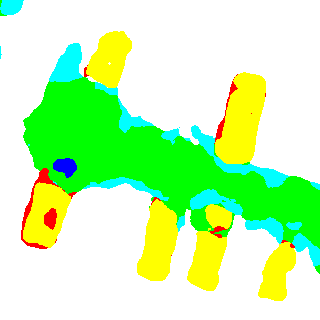}}
    {\includegraphics[width=1.5cm]{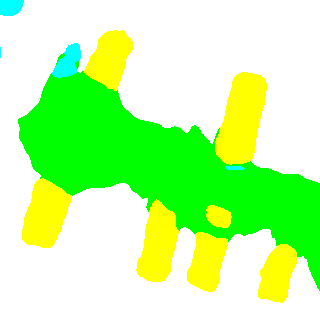}}\\
    {\includegraphics[width=1.5cm]{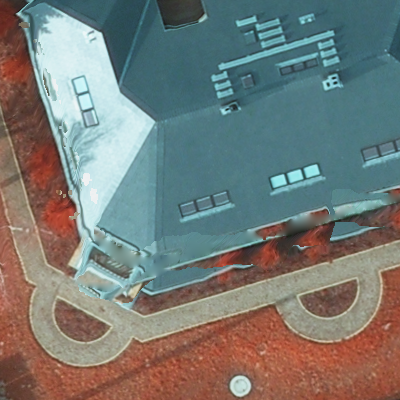}}
    {\includegraphics[width=1.5cm]{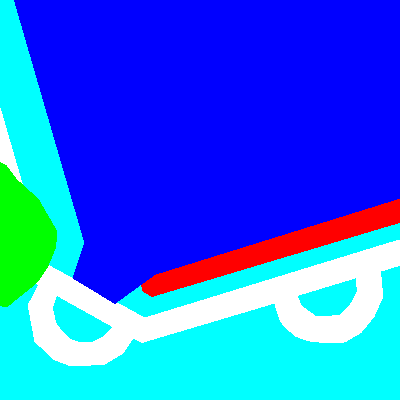}}
    {\includegraphics[width=1.5cm]{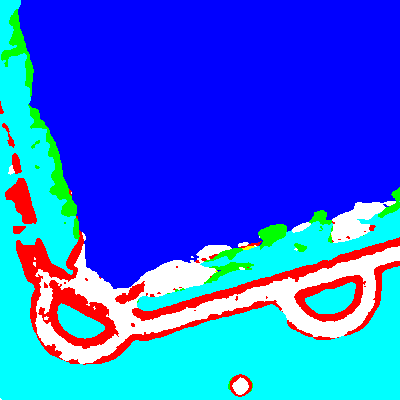}}
    {\includegraphics[width=1.5cm]{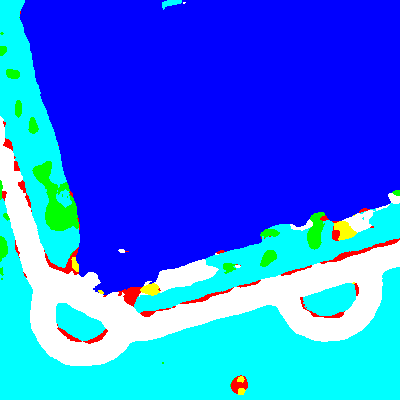}}
    {\includegraphics[width=1.5cm]{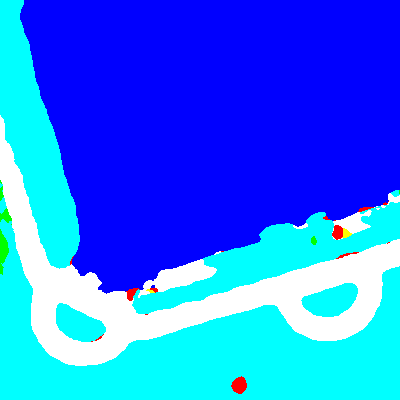}}\\
    {\includegraphics[width=1.5cm]{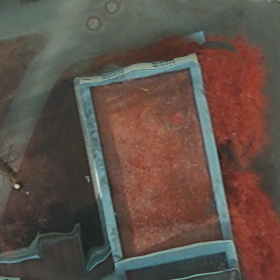}}
    {\includegraphics[width=1.5cm]{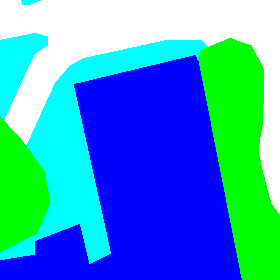}}
    {\includegraphics[width=1.5cm]{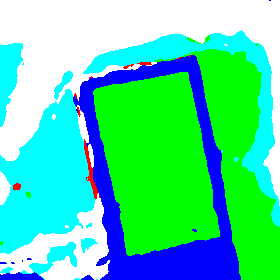}}
    {\includegraphics[width=1.5cm]{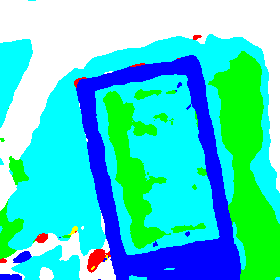}}
    {\includegraphics[width=1.5cm]{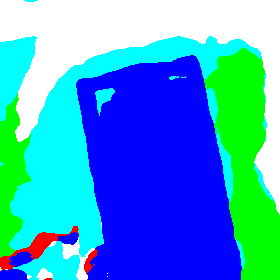}}\\
    \footnotesize
    \stackunder[5pt]{\includegraphics[width=1.5cm]{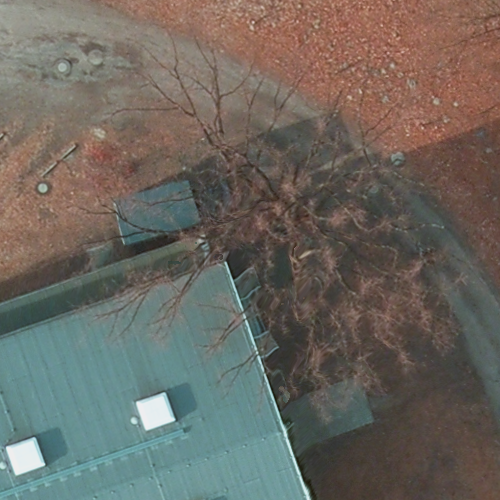}}{Image(IRRG)}
    \stackunder[5pt]{\includegraphics[width=1.5cm]{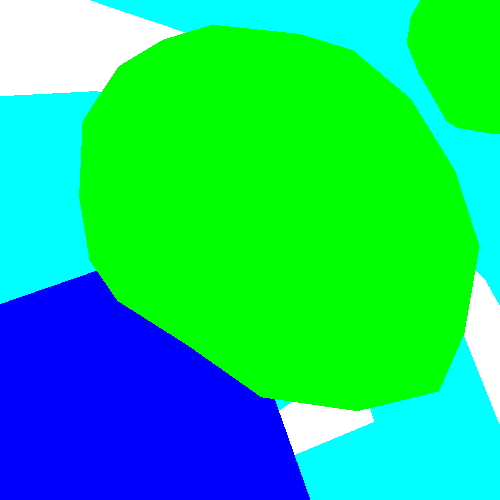}}{Reference}
    \stackunder[5pt]{\includegraphics[width=1.5cm]{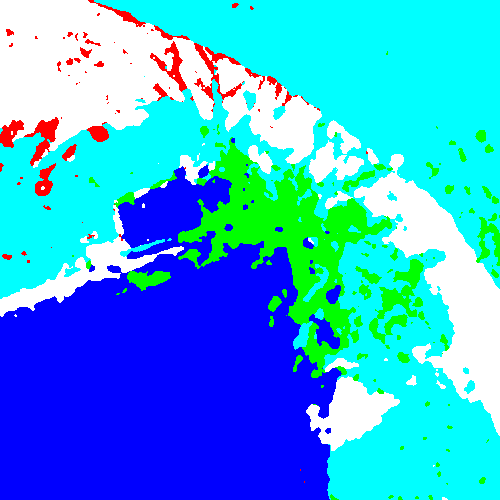}}{Before}
    \stackunder[5pt]{\includegraphics[width=1.5cm]{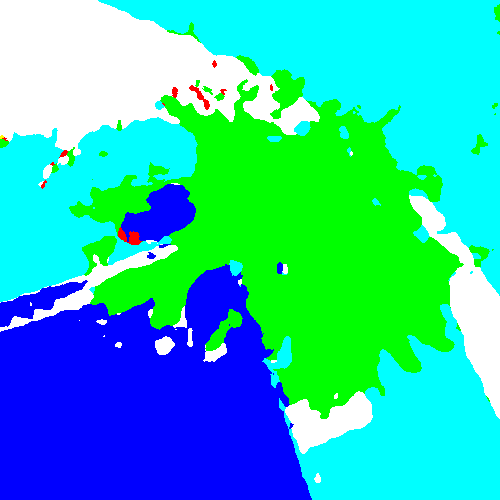}}{After PAM}
    \stackunder[5pt]{\includegraphics[width=1.5cm]{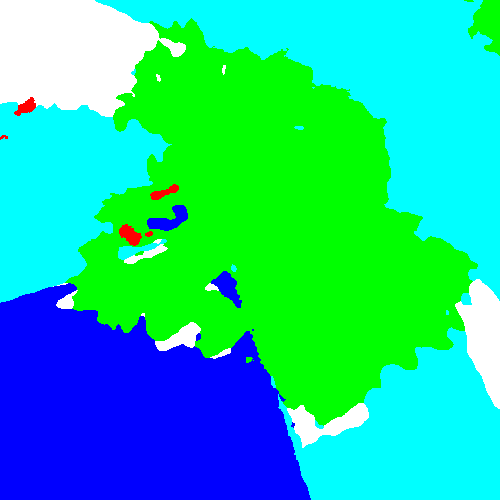}}{After AEM}
    \caption{Comparison of classified low-level features before and after the use of PAM and AEM (the Potddam dataset).}\label{Reslt_change_low}
\end{figure}

Fig.~\ref{Reslt_change_low} shows changes of the classified low-level features before and after the use of PAM and AEM. In the original low-level feature maps, pixels are only related to their neighborhoods due to the limitation of small receptive field. This leads to fragmented results and confusion of object class. However, after enhancement obtained with the proposed modules, the semantic representation of low-level features are significantly improved. The pixels are classified based on not only the surface type of objects but also the context information. Moreover, one can verify from the clearly classified boundaries that the spatial details of low-level features are kept.

\begin{table*}[htbp]
\centering
    \caption{Results on the Potsdam dataset. Per-class F1 score, average F1 score and overall accuracy (OA) are listed (\%).}
   \resizebox{1\linewidth}{!}{%
    \begin{tabular}{l|ccccc|cc}
    \toprule
   \multirow{2}*{Method} & \multicolumn{5}{c|}{Per-class F1 Score} & \multirow{2}*{average F1} & \multirow{2}*{OA}\\
   \cline{2-6}
   & Impervious Surface & Building & low vegetation & Tree & Car & \\
    \hline
    FCN & 91.46 & 96.63 & 85.99 & 86.94 & 82.28 & 88.66 & 89.42\\
    FCN+SE \cite{hu2018squeeze} & 91.47 & 96.57 & 86.21 & 87.51 & 81.07 & 88.56 & 89.55 \\
    FCN+BAM \cite{park2018bam} & 90.43 & 94.97 & 85.84 & 87.47 & 85.63 & 88.87 & 88.83\\
    FCN+CBAM \cite{woo2018cbam} & 91.37 & 96.49 & 86.00 & 87.40 & 83.22 & 88.89 & 89.46\\
    FCN+GloRe \cite{chen2019glore} & 91.55 & 96.54 & 86.17 & 87.42 & 82.69 & 88.87 & 89.57 \\
    DANet \cite{nam2017dual} & 91.61 & 96.44 & 86.11 & \textbf{88.04} & 83.54 & 89.14 & 89.72\\
    PSPNet \cite{he2015spatial} & 91.61 & 96.30 & 86.41 & 86.84 & 91.38 & 90.51 & 89.45 \\
    DeepLabv3+ \cite{chen2018deeplabv3+} & 92.35 & 96.77 & 85.22 & 86.79 & 93.58 & 90.94 & 89.74\\
    \hline
    LANet & \textbf{93.05} & \textbf{97.19} & \textbf{87.30} & \textbf{88.04} & \textbf{94.19} & \textbf{91.95} & \textbf{90.84}\\
    \bottomrule
    \end{tabular}}
        \label{ResltPD}
\end{table*}

\begin{table*}[htbp]
\centering
    \caption{Results on the Vaihingen dataset. Per-class F1 score, average F1 score and overall accuracy (OA) are listed (\%).}
       \resizebox{1\linewidth}{!}{%
    \begin{tabular}{l|ccccc|cc}
    \toprule
   \multirow{2}*{Method} & \multicolumn{5}{c|}{Per-class F1 Score} & \multirow{2}*{average F1} & \multirow{2}*{OA}\\
   \cline{2-6}
   & Impervious Surface & Building & low vegetation & Tree & Car & \\
    \hline
    FCN & 94.10 & 90.98 & 81.25 & 87.58 & 76.80 & 86.14 & 88.66\\
    FCN+SE \cite{hu2018squeeze} & 93.95 & 90.43 & 81.33 & 87.50 & 63.33 & 83.31 & 88.27\\
    FCN+BAM \cite{park2018bam} & 94.01 & 90.77 & 81.54 & 87.78 & 71.76 & 85.17 & 88.62\\
    FCN+CBAM \cite{woo2018cbam} & 94.03 & 90.86 & 81.16 & 87.63 & 76.26 & 85.99 & 88.61\\
    FCN+GloRe \cite{chen2019glore} & 93.99 & 90.57 & 81.28 & 87.49 & 70.09 & 84.68 & 88.41 \\
    DANet \cite{nam2017dual} & 94.11 & 90.78 & 81.40 & 87.42 & 75.85 &85.91 & 88.59\\
    PSPNet \cite{he2015spatial} & 94.38 & 91.44 & 81.52 & 87.91 & 78.02 & 86.65 & 88.99 \\
    DeepLabv3+ \cite{chen2018deeplabv3+} & 94.34 & 91.35 & 81.32 & 87.84 & 78.14 & 86.60 & 88.91\\
    \hline
    LANet & \textbf{94.90} & \textbf{92.41} & \textbf{82.89} & \textbf{88.92} & \textbf{81.31} & \textbf{88.09} & \textbf{89.83}\\
    \bottomrule
    \end{tabular}}
        \label{ResltVH}
\end{table*}

\noindent \textbf{Comparison with State-of-the-Art Methods.}
Comparisons are made between the proposed LANet and state-of-the-art approaches presented in literatures. All the tested approaches use the same backbone network (resnet50) and conduct the prediction on full-size test data. The experiments cover several recent works with the use of attention mechanism, including SE block \cite{hu2018squeeze}, BAM \cite{park2018bam}, CBAM\cite{woo2018cbam}  GloRe \cite{chen2019glore} and DANet \cite{nam2017dual}. The PSPNet \cite{he2015spatial} and DeepLabv3+ \cite{chen2018deeplabv3+} with receptive-field-enlarging designs are also compared. Table~\ref{ResltPD} and Table~\ref{ResltVH} reports the quantitative results on the Potsdam dataset and the Vaihingen dataset, respectively. Compared with the baseline FCN, the use of most attention-based modules such as SE, BAM and CBAM do not lead to noticeable performance improvement. The use of SE-block even causes decreases in F1 scores, especially for the car class. This is because the channel-wise descriptors are calculated on the whole feature map, and the classes that account for a small portion of total pixels are suppressed. This proves our assumption that the global-level calculation of attention descriptors is not suitable for processing large-size aerial images. The DANet with a spatial dependency modelling design improves the OA of 0.3$\%$ on the Potsdam dataset, but there is a decrease of OA on the Vaihingen dataset. DeepLabv3+, which uses both low-level features and dilated convolutions, has good performance in F1 scores. The proposed LANet outperforms existing approaches in terms of both average F1 score and OA, and has a leading in the F1 scores of all the categories.

\begin{figure*}[thpb]
        {\includegraphics[height=0.5cm]{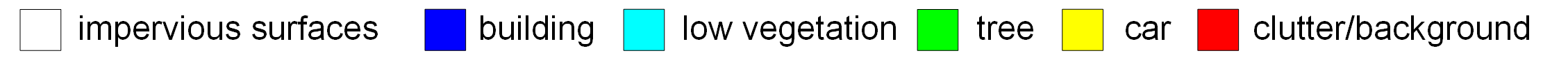}}\\
    \centering
        {\includegraphics[width=2cm]{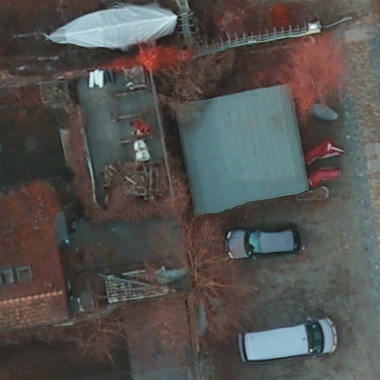}}
        {\includegraphics[width=2cm]{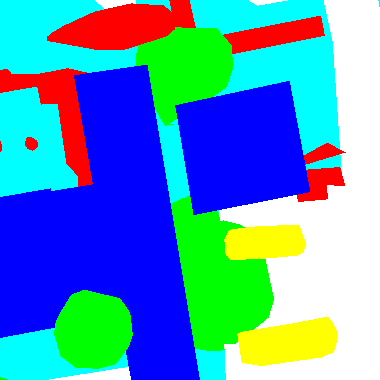}}
        {\includegraphics[width=2cm]{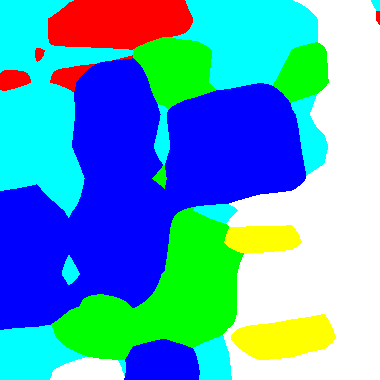}}
        {\includegraphics[width=2cm]{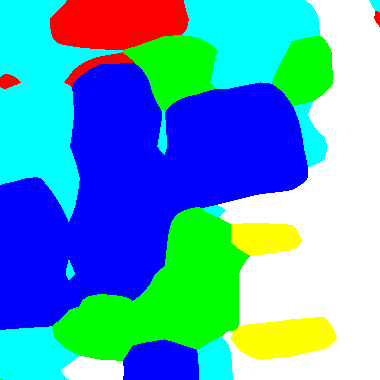}}
        {\includegraphics[width=2cm]{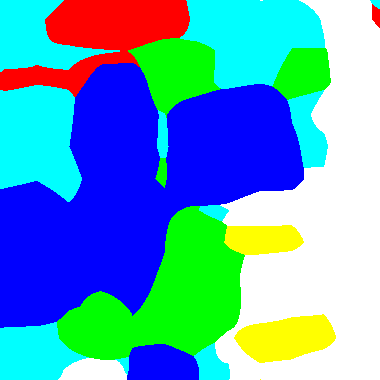}}
        {\includegraphics[width=2cm]{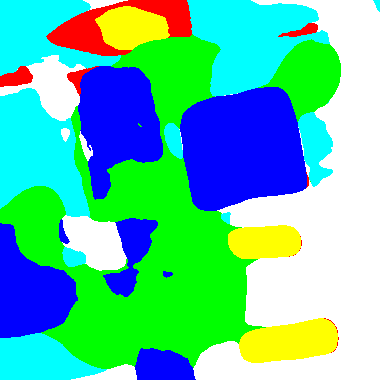}}
        {\includegraphics[width=2cm]{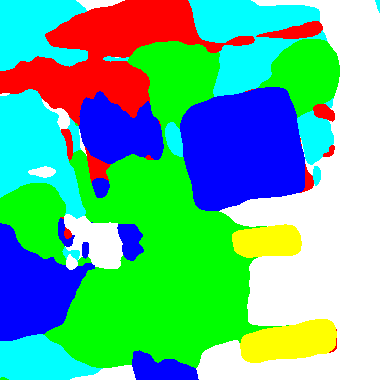}}
        {\includegraphics[width=2cm]{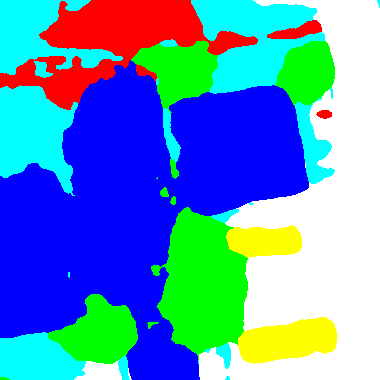}}\\
        \footnotesize
        \stackunder[5pt]{\includegraphics[width=2cm]{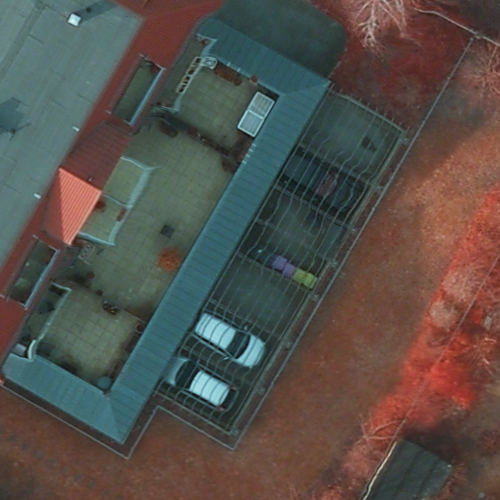}}{Image (IRRG)}
        \stackunder[5pt]{\includegraphics[width=2cm]{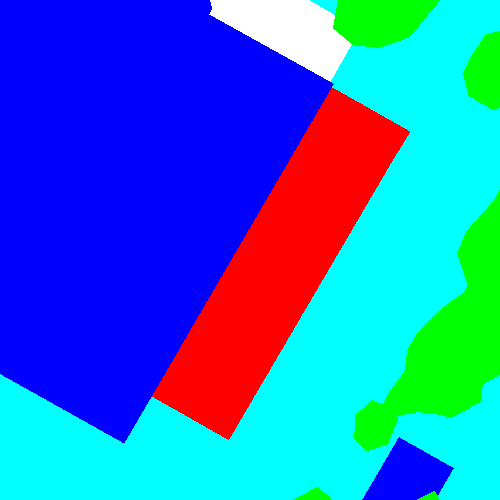}}{Ground truth}
        \stackunder[5pt]{\includegraphics[width=2cm]{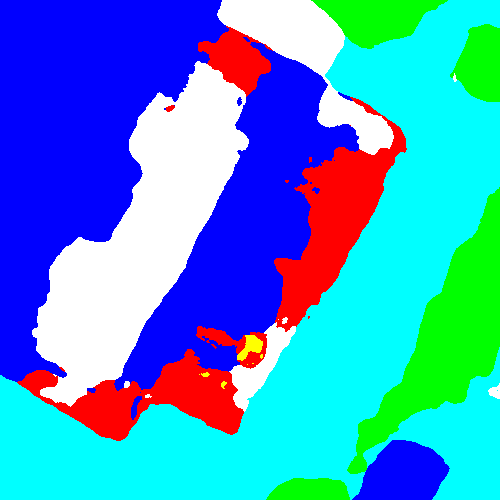}}{FCN}
        \stackunder[5pt]{\includegraphics[width=2cm]{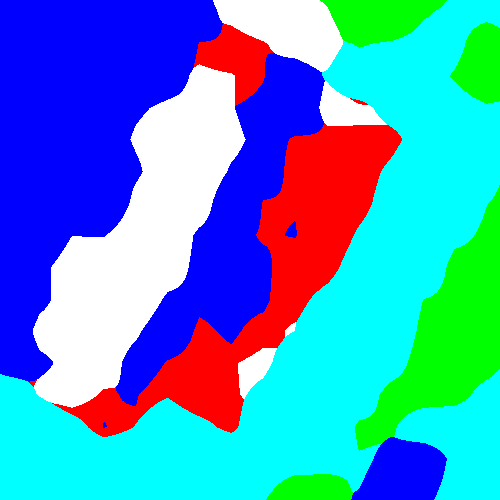}}{FCN+SE}
        \stackunder[5pt]{\includegraphics[width=2cm]{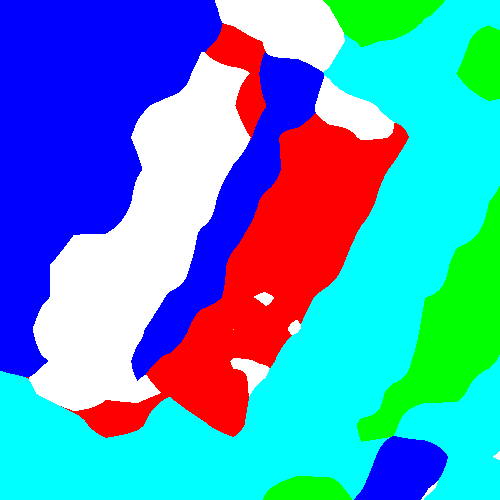}}{DANet}
        \stackunder[5pt]{\includegraphics[width=2cm]{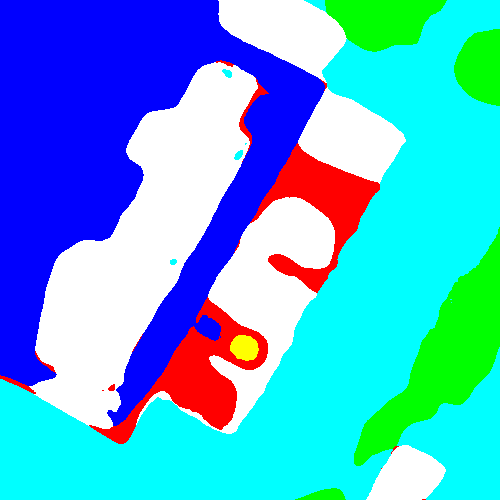}}{PSPNet}
        \stackunder[5pt]{\includegraphics[width=2cm]{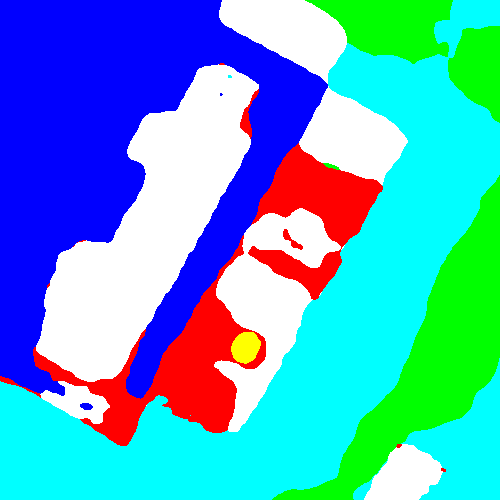}}{DeepLabv3+}
        \stackunder[5pt]{\includegraphics[width=2cm]{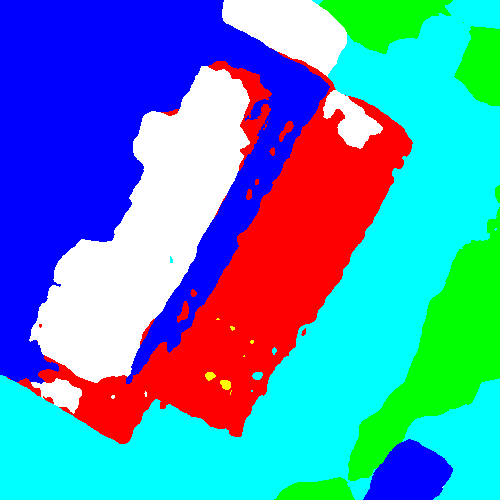}}{LANet}
    \caption{Examples of semantic segmentation results on the Potsdam dataset.}\label{detailsPD}
\end{figure*}

\begin{figure*}[thpb]
        {\includegraphics[height=0.5cm]{ColorBar_PD_wide.png}}\\
    \centering
        {\includegraphics[width=2cm]{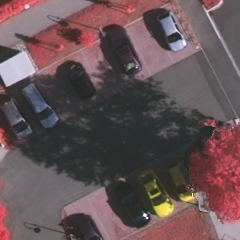}}
        {\includegraphics[width=2cm]{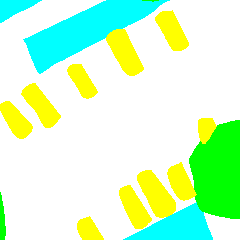}}
        {\includegraphics[width=2cm]{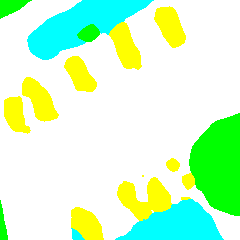}}
        {\includegraphics[width=2cm]{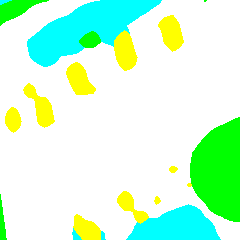}}
        {\includegraphics[width=2cm]{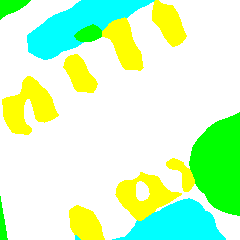}}
        {\includegraphics[width=2cm]{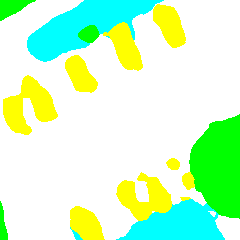}}
        {\includegraphics[width=2cm]{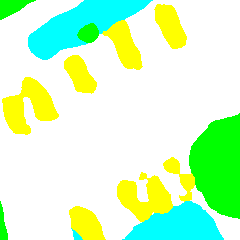}}
        {\includegraphics[width=2cm]{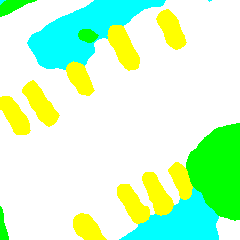}}\\
        \footnotesize
        \stackunder[5pt]{\includegraphics[width=2cm]{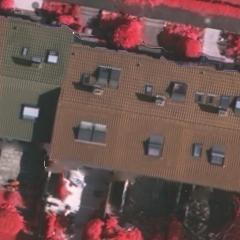}}{Image (IRRG)}
        \stackunder[5pt]{\includegraphics[width=2cm]{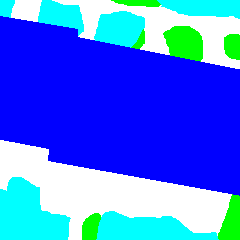}}{Ground truth}
        \stackunder[5pt]{\includegraphics[width=2cm]{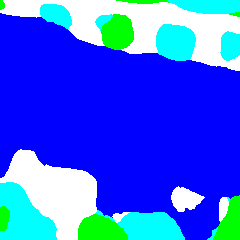}}{FCN}
        \stackunder[5pt]{\includegraphics[width=2cm]{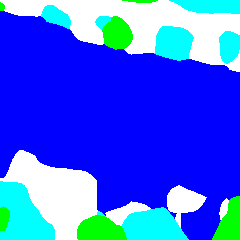}}{FCN+SE}
        \stackunder[5pt]{\includegraphics[width=2cm]{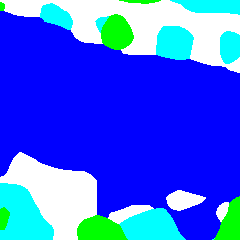}}{DANet}
        \stackunder[5pt]{\includegraphics[width=2cm]{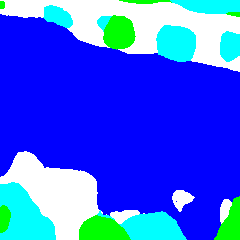}}{PSPNet}
        \stackunder[5pt]{\includegraphics[width=2cm]{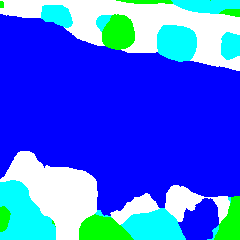}}{DeepLabv3+}
        \stackunder[5pt]{\includegraphics[width=2cm]{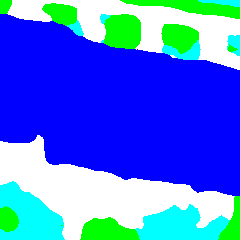}}{LANet}
    \caption{Examples of semantic segmentation results on the Vaihingen dataset.}\label{detailsVH}
\end{figure*}

\begin{figure*}[thpb]
        {\includegraphics[height=0.5cm]{ColorBar_PD_wide.png}}\\
    \centering
        \footnotesize
        \stackunder[5pt]{\includegraphics[width=4cm]{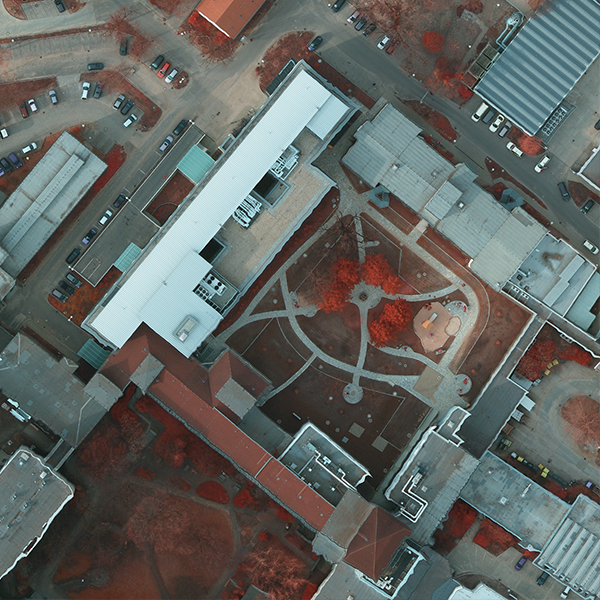}}{Test image (IRRG)}
        \stackunder[5pt]{\includegraphics[width=4cm]{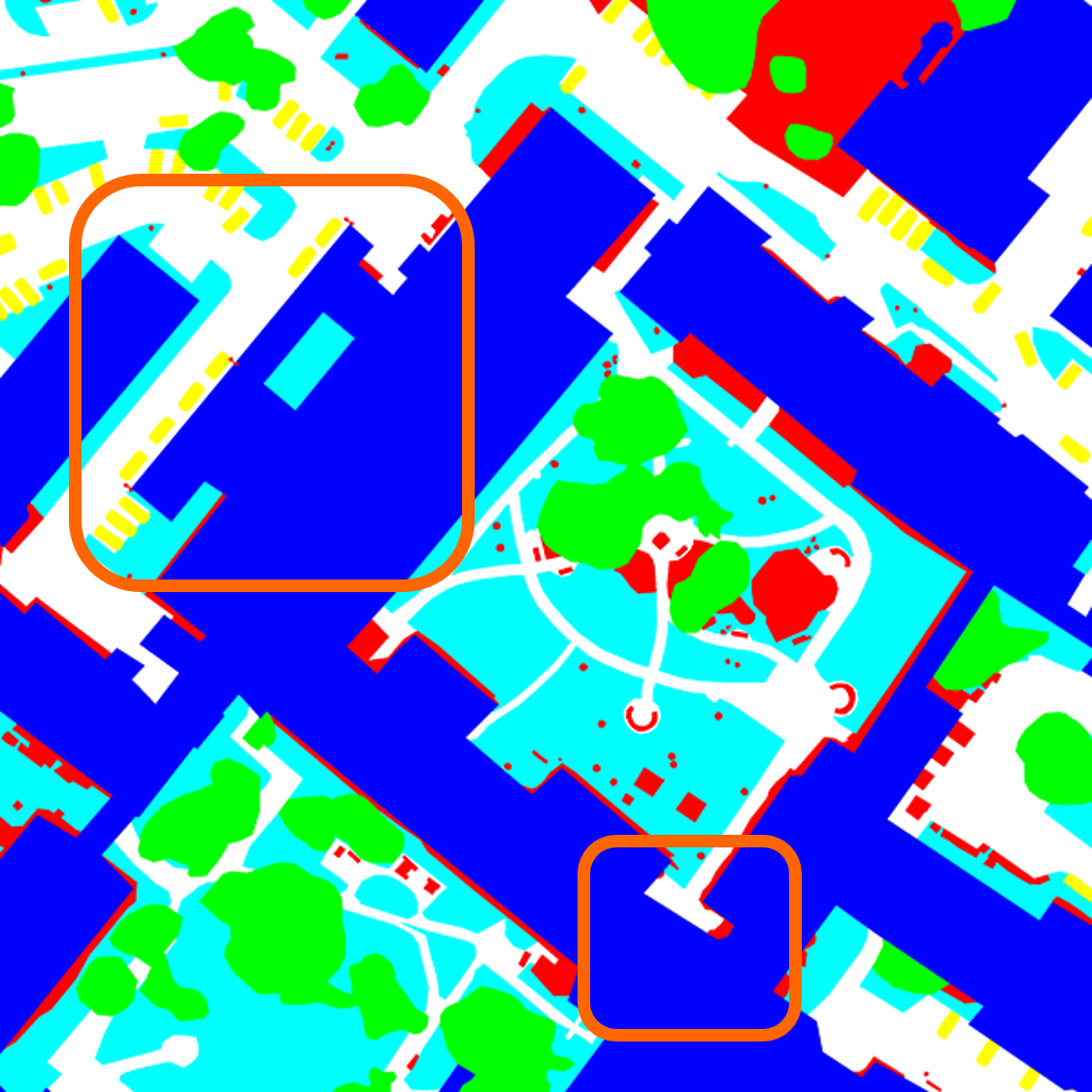}}{Ground truth}
        \stackunder[5pt]{\includegraphics[width=4cm]{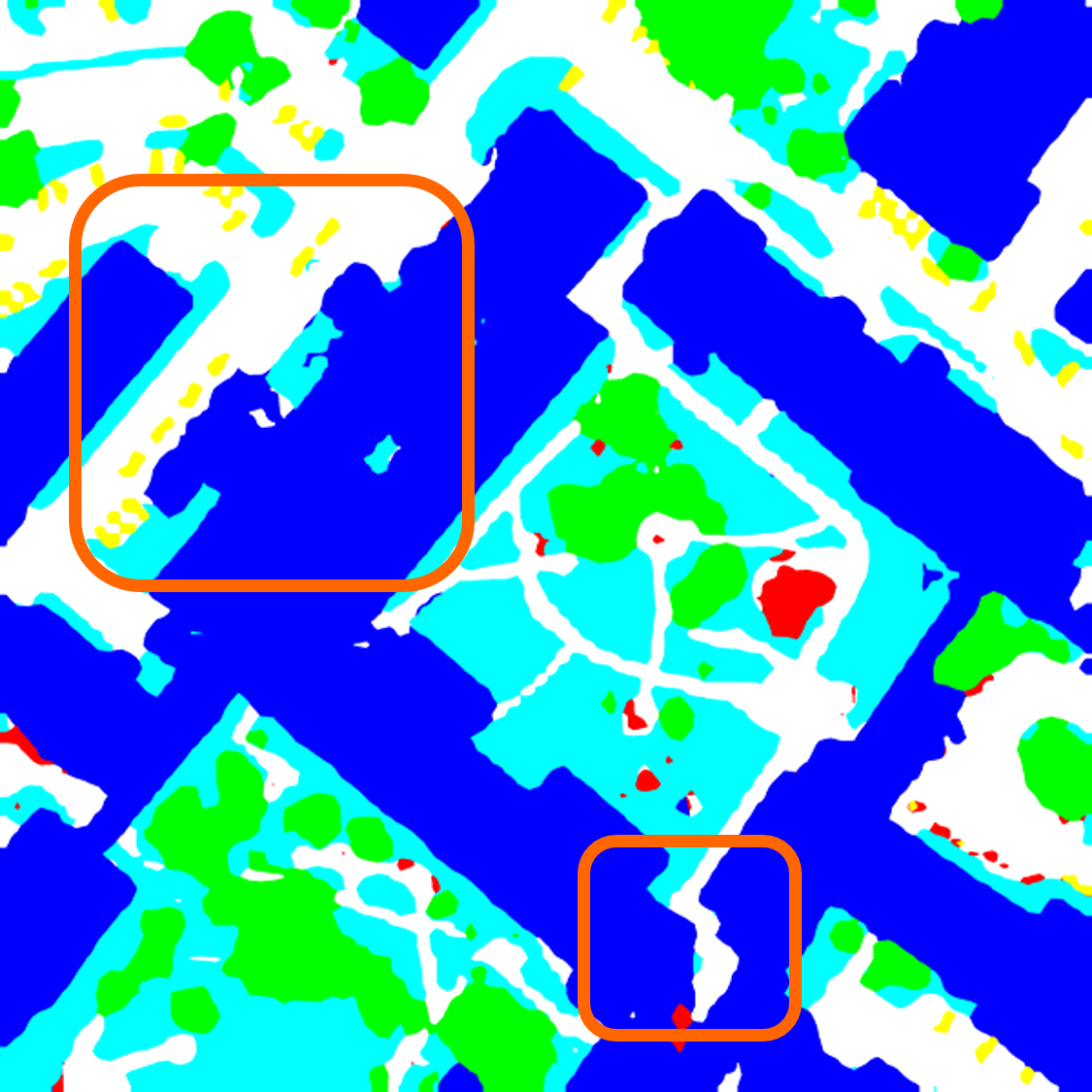}}{FCN}
        \stackunder[5pt]{\includegraphics[width=4cm]{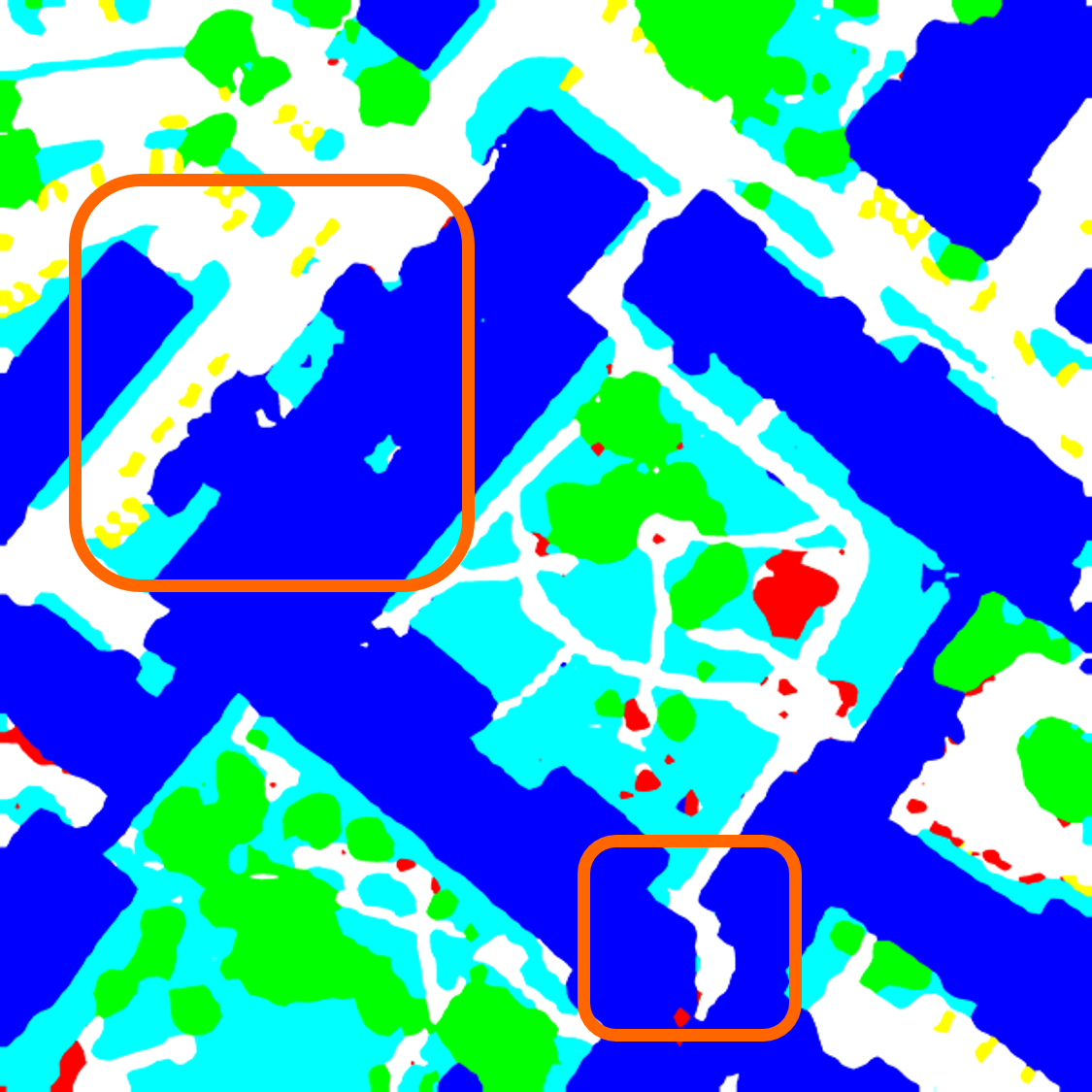}}{FCN+SE}\\
        \footnotesize
        \stackunder[5pt]{\includegraphics[width=4cm]{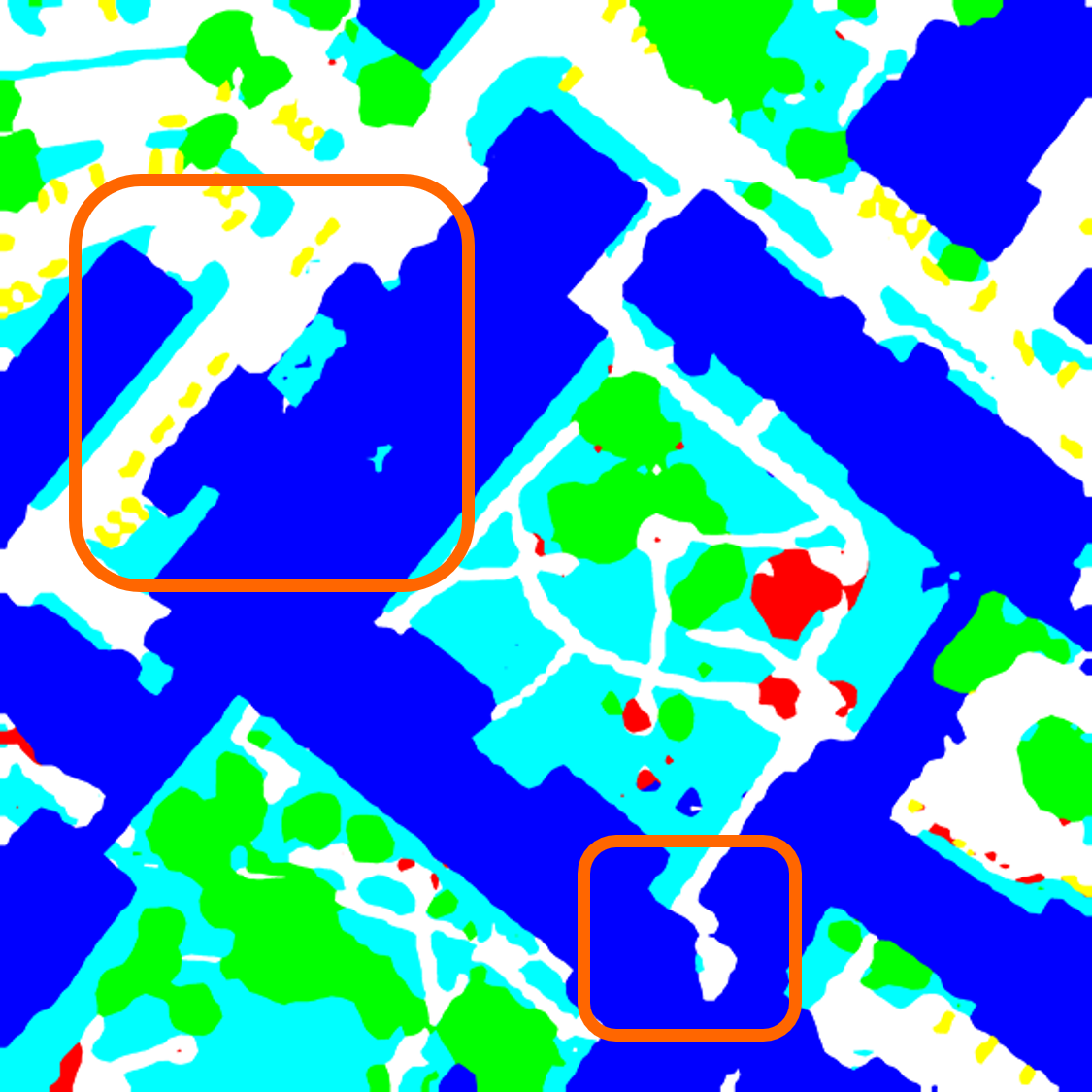}}{DANet}
        \stackunder[5pt]{\includegraphics[width=4cm]{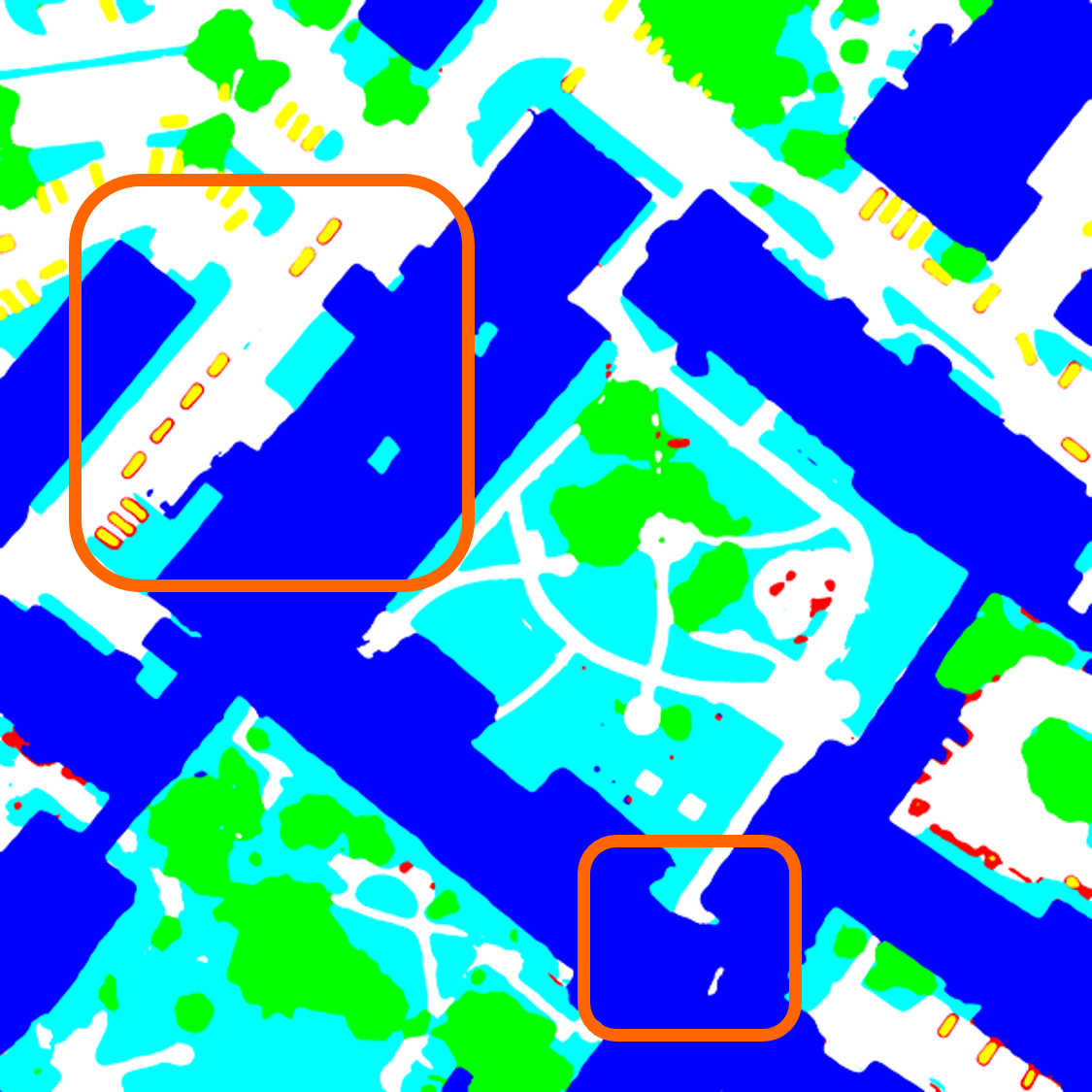}}{PSPNet}
        \stackunder[5pt]{\includegraphics[width=4cm]{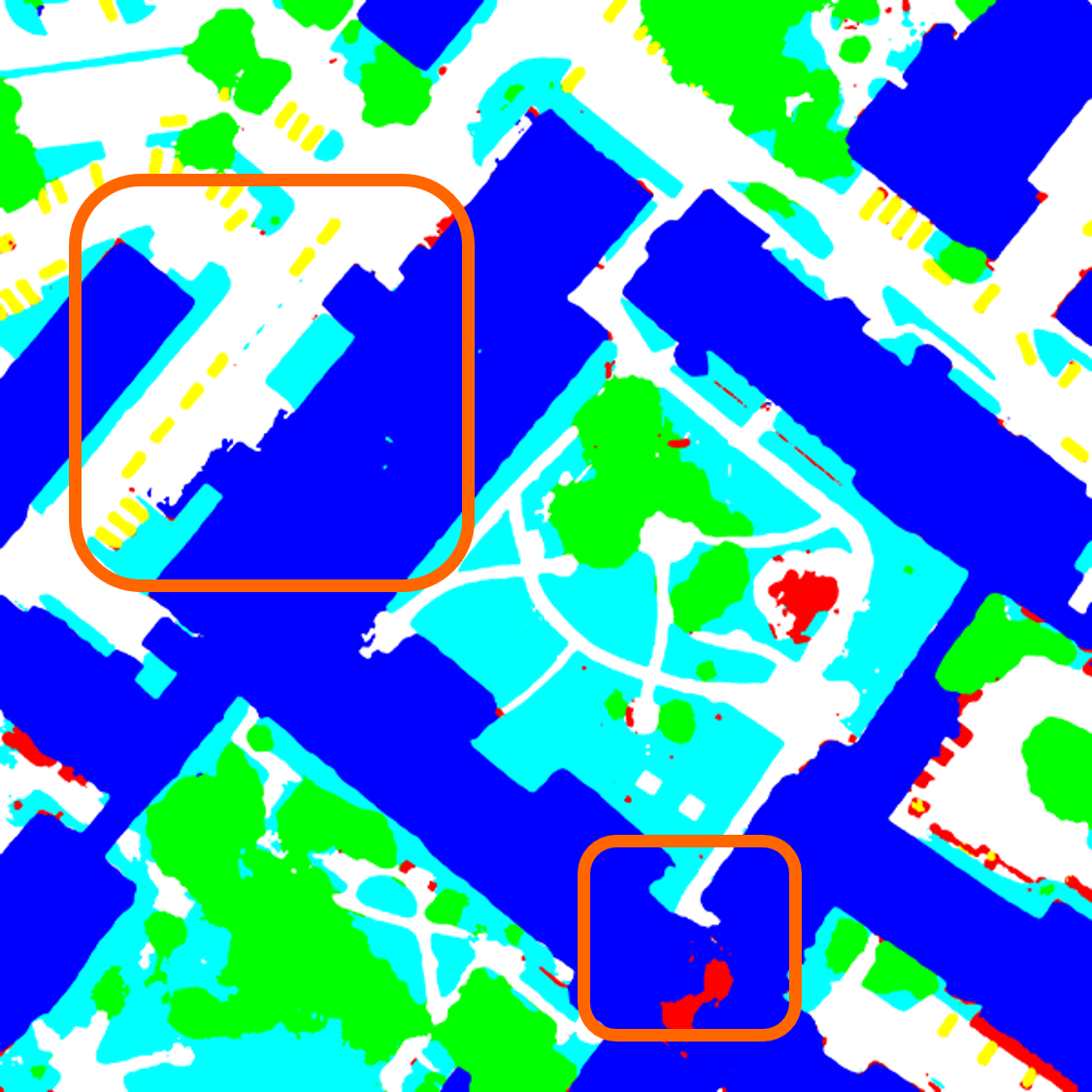}}{DeepLabv3+}
        \stackunder[5pt]{\includegraphics[width=4.1cm]{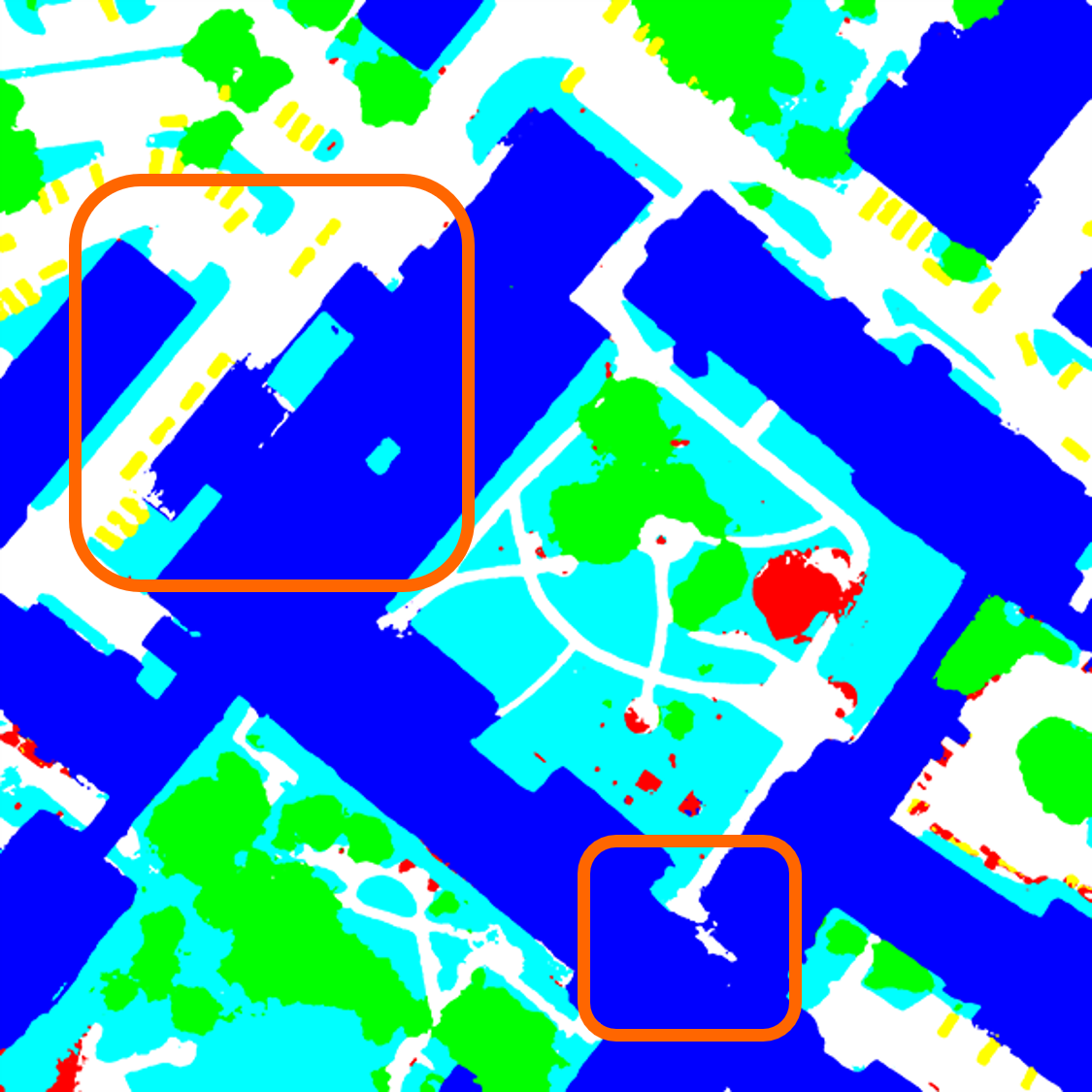}}{LANet}
    \caption{Example of large-size segmentation results on the Potsdam dataset. Major differences are highlighted.}\label{mapsPD}
\end{figure*}

\noindent \textbf{Visual Analysis of the Results.}
 Some examples of the predicted patches on the two datasets are shown in Fig.~\ref{detailsPD} and Fig.~\ref{detailsVH}, respectively. With the aggregation of local context information, results of the proposed LANet are less fragmented, while the contours of some small objects are more clear. Fig.~\ref{mapsPD} shows an example of large-size prediction result on the Potsdam dataset. The proposed LANet obtains good classification results for both large objects (e.g. buildings) and small objects (e.g. cars, paths).

\section{Conclusions}\label{sc5}
In this paper, we have presented a local attention network (LANet) to improve semantic segmentation of aerial images. Two modules are proposed for enhancing the representation of features based on the attention mechanism. Specifically, patch attention module (PAM) enhances encoding of context information based on patch-wise calculation of local descriptors, attention embedding module (AEM) embeds attention from high-level layers into low-level ones to enrich their semantic information. Experimental results on two aerial datasets (Potsdam dataset and Vaihingen dataset) show that the proposed approach greatly improves the representation of extracted features and outperform other global-attention and receptive-field-enlarging based techniques. However, there is still room for improving the encoding and enhancement of high-level features, which left for future work.

{\small
\bibliographystyle{ieee_fullname}
\bibliography{main}
}

\end{document}